\documentclass[conference]{IEEEtran}
\usepackage{amsfonts}
\usepackage{array}
\usepackage{amssymb}
\usepackage{amsmath}
\usepackage{graphicx}
\usepackage{amsfonts}
\usepackage{array}
\usepackage{color}
\usepackage[]{cite}
\usepackage{amssymb}
\usepackage{tabularx}
\usepackage{mathrsfs}
\usepackage{amsthm}
\usepackage{multirow}
\usepackage{bpchem}
\usepackage{subfigure}
\usepackage{float}
\usepackage{caption}
\usepackage{balance}

\ifCLASSINFOpdf
\else
\fi
\usepackage{amsfonts}
\usepackage{array}
\usepackage{amssymb}
\usepackage{amsmath}
\usepackage{graphicx}
\usepackage{amsfonts}
\usepackage{array}
\usepackage{color}
\usepackage[]{cite}
\usepackage{amssymb}
\usepackage{tabularx}
\usepackage{mathrsfs}
\usepackage{amsthm}
\usepackage{multirow}
\usepackage{bpchem}
\usepackage{subfigure}
\usepackage{float}
\usepackage{caption}
\begin{document}
%
\title{See the Near Future: A Short-Term Predictive Methodology to Traffic Load in ITS}

\author{\IEEEauthorblockN{Xun Zhou$^1$, Changle Li$^{1,2,*}$, Zhe Liu$^1$, Tom H. Luan$^3$, Zhifang Miao$^1$, Lina Zhu$^1$, and Lei Xiong$^2$}
\IEEEauthorblockA{$^1$State Key Laboratory of Integrated Services Networks, Xidian University, Xi'an, Shaanxi, 710071 China}
\IEEEauthorblockA{$^2$State Key Laboratory of Rail Traffic Control and Safety, Beijing Jiaotong University, Beijing, 100044 China}
\IEEEauthorblockA{$^3$School of Information Technology, Deakin University, Melbourne, VIC, 3125 Australia}
$^*${clli@mail.xidian.edu.cn}}


%


\maketitle

\begin{abstract}
The Intelligent Transportation System (ITS) targets to a coordinated traffic system by applying the advanced wireless communication technologies for road traffic scheduling. Towards an accurate road traffic control, the short-term traffic forecasting to predict the road traffic at the particular site in a short period is often useful and important.  In existing works, Seasonal Autoregressive Integrated Moving Average (SARIMA) model is a popular approach. The scheme however encounters two challenges: 1) the analysis on related data is insufficient whereas some important features of data may be neglected; and 2) with data presenting different features, it is unlikely to have one predictive model that can fit all situations. To tackle above issues, in this work, we develop a hybrid model to improve accuracy of SARIMA. In specific, we first explore the autocorrelation and distribution features existed in traffic flow to revise structure of the time series model. Based on the Gaussian distribution of traffic flow, a hybrid model with a Bayesian learning algorithm is developed which can effectively expand the application scenarios of SARIMA. We show the efficiency and accuracy of our proposal using both analysis and experimental studies. Using the real-world trace data, we show that the proposed predicting approach can achieve satisfactory performance in practice.
\end{abstract}


%
\IEEEpeerreviewmaketitle

\section{Introduction}
To predict the futuristic road traffic at a particular site is fundamental to plenty of Intelligent Transportation System (ITS) applications\cite{19,20,21,22,23,24}, such as traffic management \cite{1}, communication resources allocation \cite{2} and road-related infotainment applications \cite{3}. For instance, B\"orjesson \cite{4} applies the Swedish official long-distance model to estimate related information about traffic flow and predict demand for High Speed Rail (HSR) in order to guide investment for construction of new HSR. Gu et al. \cite{5} present that the short-term and very short-term traffic load forecasting are essential to the commitment scheduling and transmission loss estimation. Vlahogianni et al. \cite{6} have summarized relevant works within three decades and indicated that the development of short-term traffic forecasting is promoting friendly applications which can both provide accurate information to drivers and be used for signal optimization. In a nutshell, to achieve accurate traffic prediction is important to the performance of many advanced ITS applications.

These existing prediction schemes are typically by exploring the correlated historical data, which can be classified as linear methods and non-linear methods based on the prediction functions adopted. The linear methods include the grey forecasting GM(1,1) \cite{7} , error component model \cite{8} and the Autoregressive Integrated Moving Average (ARIMA) Model \cite{9} with fixed prediction functions which explores the assumption of linearity and stationarity of the prediction function to infer future transportation trends. Compared with the linear methods which only adjusts related coefficients according to historical data, the non-linear methods can better learn data features from huge samples and therefore build adaptive prediction functions. Corresponding prediction model can describe non-linear characteristics and achieve more accurate forecasting performance in transportation systems. Currently typical non-linear methods such as machine learning models \cite{10, 11, 13}, have been applied in several application fields from traffic forecasting to communication allocating \cite{13}.


However, in the short-term traffic forecasting field, SARIMA \cite{14} which is improved from ARIMA is also a popular linear method which can outperform many non-linear methods. In contrast with most learning models, Marco et al. \cite{13} prove that the typical SARIMA coupled with a Kalman filter would work better in traffic forecasting than other learning models under the same conditions. There are two major reasons. Firstly, SARIMA based on regression analysis possesses the machine learning features, which is effective for applying history data to establish prediction model. Secondly, with respect to most non-linear model applying history data to obtain a global optimum result, SARIM can utilize periodicity existing in data to prevent uncorrelated samples influencing the performance of the model. Considering the remarkable periodicity consisting in traffic flow, SARIMA is the most appropriate model for traffic forecasting.

SARIMA encounters two challenges, which will be investigated in this work. The first one is that related data used in the models usually lacks analysis. Beforehand analyzing traffic data features can be useful for adapting SARIMA to traffic flow. The second one is that different data sensitivity makes it unreasonable to apply only one predictive model in all situations. Although there are some fixed hybrid models, however, employing machine learning to combine models is advantageous to revised the hybrid structure dynamically and reduce the effect from unexpected incidents.

We address the above challenges in two aspects. First of all, we extract stable part of traffic flow as a constant, i.e., the traffic flow constant, and the remainder is the fluctuant part which is proved to follow Gaussian distribution. Combining the autocorrelation property in traffic flow, we revise the structure of SARIMA only to predict fluctuant part. Furthermore, applying the Gaussian distribution feature, a hybrid model is proposed based on the revised SARIMA to adjust forecasted results dynamically. In particular, we highlight our main contributions in this paper as follows:

\begin{itemize}
  \item Based on the analysis of traffic flow, we define the traffic flow constant and calculate residuals among real data and constant as fluctuant part to revise the structure of original. Accuracy of it can be improved.
  \item We establish a Bayesian learning algorithm to combine with classical SARIMA and revised SARIAM to obtain a hybrid model to expand its application range and improve its stability.
  \item Real data is used to verify the performance improvement of hybrid model and thoroughly analyze this model.
\end{itemize}

The rest of this paper is organized as follows. Section II introduces preliminary works, and displays the process of data preprocessing.  Section III is about detailed information of revised process and hybrid model. Furthermore, in Section IV, simulation results are discussed to examine the performance of hybrid model based on real data of subway traffic flow. Finally, Section V concludes this paper and future works.
\section{Preliminary}
This part is divided into two subsections. In the first one, we summarize the SARIMA processes in order to introduce the notations used in the remainder of the paper. In the second one, we also reveal relevant information about our data used in the simulation. It is a common part in the literatures about data processes. Although this part may be seen as non-technical, however, it is important for coming works.
\subsection{SARIMA Model}
SARIMA model was proposed by George et al. based on ARIMA. It is skilled at tackling time series that exhibits an $s$-periodic behavior. The $s$ denotes that similarities in the series occur after $s$ basic time intervals. For example, the seasonality existing in daily models \cite{15} is 5 days. The model is often shown as SARIMA($p,d,q$)($P,D,Q{)_s}$ and the function is
\begin{eqnarray}
{\phi _p}(B){\varphi _P}({B^s}){\nabla ^d}\nabla _s^D{z_t} = {\theta _q}(B){\Theta _Q}({B^s}){a_t},
\end{eqnarray}
where ${a_t}$ is a white noise. $B$ is the backward-shift operator, i.e., $B{z_t} = {z_{t - 1}}$. $\nabla$ is the differencing operator, i.e., $\nabla  = 1 - B$. $\phi (B)$, $\varphi ({B^s})$, $\theta (B)$ and $\Theta (B^s)$ are the polynomials in $B$ and ${B^s}$ respectively. $p, d, q, P, D$, and $Q$ are degrees of corresponding polynomials, i.e.,
\begin{eqnarray}
{\phi _p}(B) = 1 - {\phi _1}{B^1} - {\phi _2}{B^2} - ... - {\phi _p}{B^p},\
\end{eqnarray}
\begin{eqnarray}
{\varphi _P}({B^s}) = 1 - {\varphi _1}{B^s} - {\varphi _2}{B^{2s}} - ... - {\varphi _P}{B^{Ps}},
\end{eqnarray}
\begin{eqnarray}
{\nabla ^d} = {(1 - B)^d},
\end{eqnarray}
\begin{eqnarray}
\nabla _s^D = {(1 - {B^s})^D}.
\end{eqnarray}
The (1) can be transformed as a more common form in (6)
 \begin{eqnarray}
\begin{array}{l}
{z_t} = {\alpha _1}{z_{t - 1}} + {\alpha _2}{z_{t - 2}} + ... + {\alpha _n}{z_{t - p - d - Ps - Ds}} + {a_t}\\
\begin{array}{*{20}{c}}
{}&{}
\end{array}{\beta _1}{a_{t - 1}} + {\beta _2}{a_{t - 2}} + ... + {\beta _m}{a_{t - q - Qs}}.
\end{array}
\end{eqnarray}
The $t - n$ denotes there are $n$ time intervals and the predicted result can be denoted as (7)
 \begin{eqnarray}
\begin{array}{l}
{\mathop z\limits^\sim}_t = {\alpha _1}{z_{t - 1}} + {\alpha _2}{z_{t - 2}} + ,..., + {\alpha _n}{z_{t - p - d - Ps - Ds}}\\
\begin{array}{*{20}{c}}
{}&{}
\end{array}{\beta _1}{a_{t - 1}} + {\beta _2}{a_{t - 2}} + ... + {\beta _m}{a_{t - q - Qs}}.
\end{array}
\end{eqnarray}
Therefore, the ${a_t}_{ - n} = {z_{t - n}} - {\mathop z\limits^\sim}_{t - n}$ is the residual between history measured data and predicted data. The  least square method is used to train samples to calculated related coefficient of each parameter.
\subsection{Data Processing}
In this paper, the real data is downloaded from the New York State Home. It is a public data source established by USA government. The more detailed information can be found on \cite{16}. Data is true and collected by devices on turnstiles, which can count the number of people entering into subway stations, and the data is often uploaded each 4 hours. Attributes such as subway stations number, time stamps and the number of entrancing people are included.

We choose the raw data sets collected from January to March in 2016 as samples. Firstly, raw data sets are classified based on the attribute, station ID. We randomly extract relevant data sets of one subway station and exclude data collected on the public holidays, weekends and bad weather, which may influence the people flow seriously, from the original data sets. Then, the data sets are cleared. Some data collected on the special devices is ignored, for example the devices are fault or unavailable, and redundant members which may be recorded for many times are deleted. As for the missing value, it is interpolated with the mean value. Data used in this paper is the number of people entering into the station at six time segments and they are respectively 03:00-07:00, 07:00-11:00, 11:00-15:00, 15:00-19:00, 19:00-23:00 and from the day 23:00 to the next day 03:00 (23:00-03:00). Finally, six data sets can be acquired. These data sets includes peak and off-peak of people flow. In this paper, data of two months is used as training sets to establish related model and Bayesian learning algorithm. After this, data of one week is used as testing sets to evaluate these models mentioned above.

\section{Bayesian Seasonal Autoregressive Integrated Moving Average}
In this part, we present our scheme, Bayesian Seasonal Autoregressive Integrated Moving Average (BSAIMA), including the theoretical analysis, model confirmation and optimization respectively.
\subsection{Model Revision}
Generally speaking, it is accessible for us to predict the traffic flow based on the hourly model in \cite{15}, whose time interval in (7) is one hour. It can be shown as (8)
\begin{eqnarray}
z(t) = f\{ z(t - 1),z(t - 2),...,z(t - n)\},
\end{eqnarray}
where $f$\{ . \} defines the forecast algorithm in (7). One hour of time interval is often the upper bound for guaranteeing the efficiency of the model. The major reason is that time interval is so great leading the information getting from the adjacent time intervals to become independent. Related information can be known from the Burke theory\cite{17} that the leaving flow is irrelevant with the number of people existing in the system.  Thus, it is necessary to establish a prediction model that is more reliable for different data sets.

In normal condition, it is reasonable to assume the people in each time segment can be defined as (9)
\begin{eqnarray}
{z_i} = {d_i} + {\varepsilon _i},
\end{eqnarray}
where ${z_i}$ denotes the traffic flow in $i$th time segment. We define ${d_i}$ in this equation as traffic flow constant which is the number of constant part consisting of such as office workers and students. ${\varepsilon _i}$ is fluctuation around the constant ${d_i}$. It is from a lot of wispy factors such as weather and ticket price. So it is reasonable to consider ${\varepsilon _i}$ as the Gaussian distribution with zero mean. All of these can be verified based on the law of large numbers. The above (9) can be converted as (10)
\begin{eqnarray}
\begin{array}{l}
\mathop {\lim }\limits_{n \to \infty } \frac{1}{n}\sum\limits_{j = 1}^n {z_i^j}  = \mathop {\lim }\limits_{n \to \infty } \frac{1}{n}\sum\limits_{j = 1}^n {(d_i^j + \varepsilon _i^j)}, \\
\mathop {\lim }\limits_{n \to \infty } \frac{1}{n}\sum\limits_{j = 1}^n {z_i^j}  = {d_i},
\end{array}
\end{eqnarray}
where $n$ denotes the number of samples used in statistical algorithm and $j$ denotes the order of samples rather than power. Based on the assumption the that expectation of ${\varepsilon _i}$ is zero, the constant ${d_i}$ can be obtained by calculating the mean of $n$ samples. In order to verified our assumption, samples collected from two months are used to calculate the ${d_i}$ and ${\varepsilon _i}$. In the next part, based on the Kolmogorov-Smirnove (K-S) test\cite{18}, we can verify the Gaussian distribution.

Now, two import conclusions can be summarized. The first one, people flow at the same time segment in different days may be relevant, because they have the same cardinal number ${d_i}$ and the random variable following the same distribution. The second one, applying the constant ${d_i}$ and discussing the variation of ${\varepsilon _i}$ can be a forecasting way. Thus, the daily model\cite{15} can be seen as more perfect choice, which time interval is one day. At the same time, assuming seasonality existing also in the relevant data is advisable.  For example, in each Monday, majority of companies and schools often will hold meeting to summarize the work of last week, so people will arrive earlier than other day and the peak will also arrive earlier. So, (11) can be used as forecast algorithm
\begin{eqnarray}
{z^n}(t) = f\{ {z^{n - 1}}(t),{z^{n - 2}}(t),...,{z^{n - j}}(t)\}.
\end{eqnarray}
It means the traffic flow in $n$th day at time $t$ can be estimated by the data collected from the other days before $n$th day. Considering the second conclusion, we can divide the traffic flow into fluctuant and stable parts. Only the fluctuant part is predicted in the (11). Then integrating predicted results and constant can obtain traffic flow in next time interval. It is a effective way by reducing the predicted content to decrease error of prediction. So, the (11) can be revised as
\begin{eqnarray}
{z^n}(t) = f\{ {\varepsilon ^{n - 1}}(t),{\varepsilon ^{n - 2}}(t),...,{\varepsilon ^{n - j}}(t)\}  + d(t),
\end{eqnarray}
 and the common form revised from (7) in the predicted task can be shown as (13)
 \begin{eqnarray}
\begin{array}{l}
{z^n}(t) = {\alpha _1}{\varepsilon ^{n - 1}}(t) + {\alpha _2}{\varepsilon ^{n - 2}}(t) + ... + {\alpha _j}{\varepsilon ^{n - j}}(t)\\
\begin{array}{*{20}{c}}
{}&{}&{}
\end{array} + {\beta _1}{a^{n - 1}}(t) + {\beta _2}{a^{n - 2}}(t) + ... \\
\begin{array}{*{20}{c}}
{}&{}&{}
\end{array}+ {\beta _m}{a^{n - m}} + d(t),
\end{array}
\end{eqnarray}
$j \in [1,p + d + Ps + Ds]$ and $m \in [1,q + Qs]$. Model described by (12) and (13) is named Residual Seasonal Autoregressive Integrated Moving Average (RARIMA).In order to determine the value of $p, d, q, P, D$, and $Q$, the least square method is used to fit minimum Mean Absolute Error (MAE) and Goodness of Fit ${R^2}$ of different order combination is checked to choose the largest one. However, in the (1), because of polynomials in $B$, the SARIMA model always includes several items such as ${B^{s - 1}}$ and ${B^{s - 2}}$. Although, ${z_t}$ is strongly correlated with the $s$th item, i.e., ${B^{s}}$ due to seasonality, in fact the adjacent items around the $s$th item is often irrelevant. It will influence the accuracy of SARIMA. A instance will be used to explain the fact.

At first, the correlation among these samples can be calculated based on the(14) proposed by George et al. \cite{9}
\begin{eqnarray}
{p_l} = \frac{{\sum\limits_{t = l + 1}^T {({z_t} - \mathop z\limits^\_ )({z_{t - l}} - \mathop z\limits^\_ )} }}{{\sum\limits_{t = 1}^T {{z_t} - \mathop z\limits^\_ } }},
\end{eqnarray}
where ${z_t}$ is the real data at time $t$, ${\mathop z\limits^\_}$ is the mean of training set, $l$ is the lag object, $T$ is the size of training set, ${p_l}$ is the value of correlation between ${z_t}$ and ${z_{t - l}}$. The value of ${p_l}$ is greater, the ${z_t}$ is more similar with ${z_{t - l}}$. Fig. 1 is about autocorrelation and partial autocorrelation of the traffic flow collected from January to March at peak 07:00 to 11:00.
\begin{figure}
  \centering
  \includegraphics[width=3in]{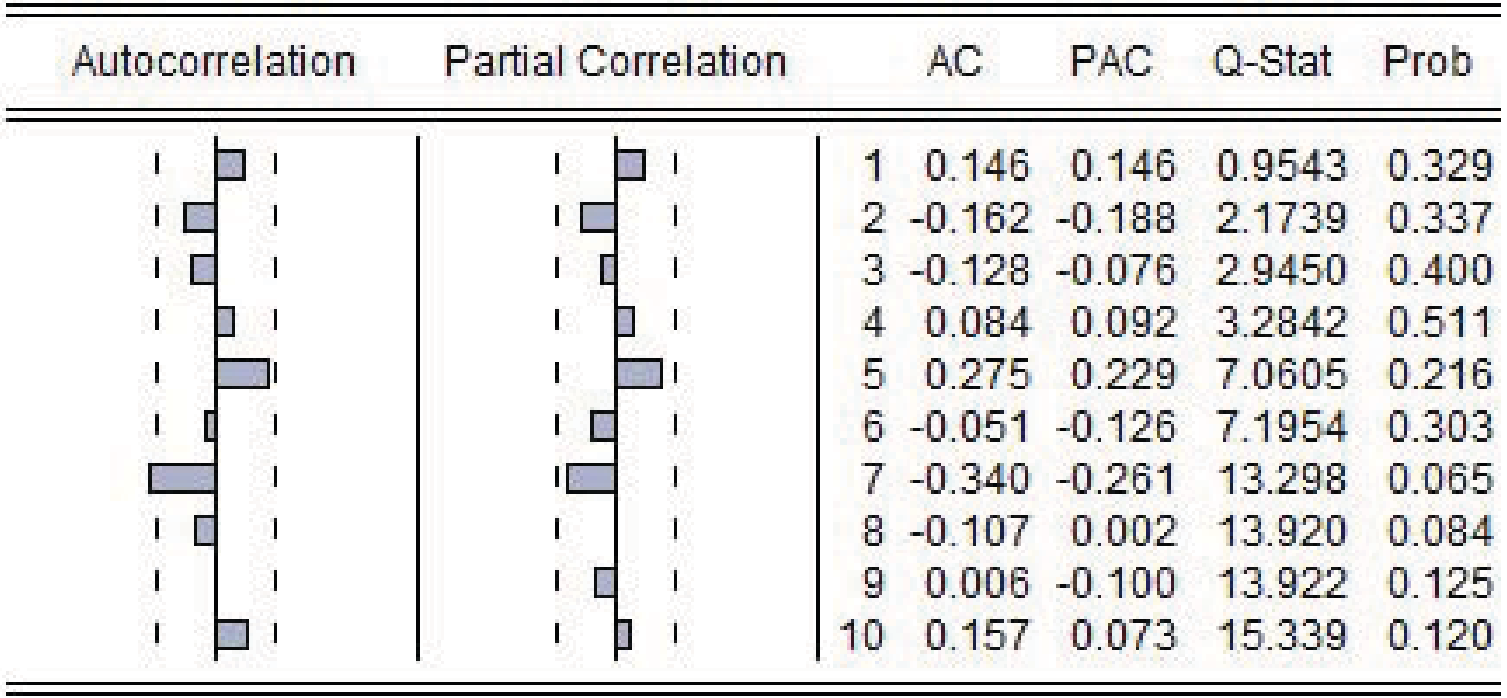}\\
  \caption{Autocorrelation and Partial Autocorrelation of the Traffic Flow Collected from January to March (7:00-11:00)}
\end{figure}
When the value of correlation is around the double standard line (dash line in the picture), it represents the current data is strongly related with the corresponding lag object\cite{9}. From the picture we can see the lag objects 5 and 7 are strongly related with the current traffic flow. The SARIMA(2,0,2)(1,0,0$)_5$ is the best model and it can be shown as
 \begin{eqnarray}
\begin{array}{l}
{z^n}(t) = {\alpha _1}{z ^{n - 1}}(t) + {\alpha _2}{z ^{n - 2}}(t)  + {\alpha _5}{z ^{n - 5}}(t)\\
\begin{array}{*{20}{c}}
{}&{}&{}
\end{array}+{\alpha _6}{z ^{n - 6}}(t) + {\alpha _7}{z ^{n - 7}}(t)\\
\begin{array}{*{20}{c}}
{}&{}&{}
\end{array}+{\beta _1}{a^{n - 1}}(t) + {\beta _2}{a^{n - 2}}(t).
\end{array}
\end{eqnarray}
In order to cover the relevant item 5 and 7. Two irrelevant items 1 and 6 is also included. At the same time, the equation contains so many lag objects which may lead to over fitting. In order to reject irrelevant items and avoid over fitting, at most three relevant items should be used in autoregression models\cite{9}. We only use the top three items on the value of correlation to establish model. The (15) can be transformed as
 \begin{eqnarray}
\begin{array}{l}
{z^n}(t) = {\alpha _2}{\varepsilon ^{n - 2}}(t) + {\alpha _5}{\varepsilon ^{n - 5}}(t) + {\alpha _7}{\varepsilon ^{n - 7}}(t)\\
\begin{array}{*{20}{c}}
{}&{}&{}
\end{array} + {\beta _2}{a^{n - 2}}(t) + {\beta _5}{a^{n - 5}}(t)\\
\begin{array}{*{20}{c}}
{}&{}&{}
\end{array} +{\beta _7}{a^{n - 7}} + d(t).
\end{array}
\end{eqnarray}
The simple form of the transformation (15) is define as S-ARIMA$(top. 1, top. 2, top. 3)$. $top. 1, top. 2$, and $top. 3$ denotes the order of the three relevant items whose the values of correlation are in the top three. So, the (16) is shown as S-ARIMA(2,7,5). The model defined in the (13) should be denoted as RARIMA(top. 1, top. 2, top. 3). Then the finally function used to describe RARIMA can be shown as (17)
 \begin{eqnarray}
\begin{array}{l}
{z^n}(t) = {\alpha _{top. 1}}{\varepsilon ^{n - top. 1}}(t) + {\alpha _{top. 2}}{\varepsilon ^{n - top. 2}}(t)\\
\begin{array}{*{20}{c}}
{}&{}&{}
\end{array}+ {\alpha _{top. 3}}{\varepsilon ^{n - top. 3}}(t) + {\beta _{top. 1}}{a^{n - top. 1}}(t)\\
\begin{array}{*{20}{c}}
{}&{}&{}
\end{array}+ {\beta _{top. 2}}{a^{n - top. 2}}(t) +{\beta _{top. 3}}{a^{n - top. 3}} + d(t).
\end{array}
\end{eqnarray}

Data sets used into the two revised models must be stable, if not it should previously perform differencing on the data sets to make it stable. Then after checking adjusted ${R^2}$ to determine the order combination of the model and the least square method is used to determine coefficient of each item to obtain minimum MAE.
\subsection{Model Optimization}
In this part, we optimize the model mentioned above. Although the inspiration actually derived from the compared results in simulation part, the theory is introduced now.

Maybe one model can be more accurate than other models on MAE of sample sets. However, it does not mean it is more excellent than others in each step. For example, the RARIMA model does well in tracking the peak about traffic flow, however, when the people flow is stable in several time segments, the S-ARIMA model is more effective than it. If we can combine the two models to predict traffic flow, it is useful to gain a more accurate result. In other words, before predicting the traffic flow, we should estimate which models will perform better.

In order to solve the question, classification function in machine learning is considered. Based on the history data, corresponding information can be used to establish relevant functions to describe features existing in the different models. Combining with the related features of models and correlate information about current data, it can be available to choose the model with best performance in this prediction. In addition, learning algorithm can perfect the relevant functions with increase of the size of history data sets, and revised the hybrid structure dynamically to reduce the effect from unexpected incidents. In this paper, Bayesian decision theory is used to generated the hybrid model, because it can rely on fewer data features to gain a desired achievement. These detailed steps are as follow:
\subsubsection{Step 1}
In the Bayesian decision theory, the information about class should be clear at first, such as the number of the class. In this paper, each model is seen as one class and show as ${C_i}$, $i \in \{ A,B,C,...\}$. In this paper only two models are combined, namely S-ARIMA and RARIMA, so they are defined as ${C_A}$ and ${C_B}$ respectively.
\subsubsection{Step 2}
Attributes about each class should be clear. It is a difficult work for us, because the type of these attributes should be same for each class at first. Then, the condition probability of these attributes should be possible for being calculated. Finally, the most important one is these attributes can distinguish each member. In this paper, the residual about predicted value and mean value of history is chosen as attribute, it can be shown as the (19)
\begin{eqnarray}
{\varepsilon _i} = {\mathop z\limits^ \sim}_i - {d_i}.
\end{eqnarray}
Since ${\varepsilon _i}$ can been seen to follow Gaussian Distribution reasonably, it is useful for us calculated the condition probability.
\subsubsection{Step 3}
The Bayesian decision theory is based on the posterior probability to determine which class the member should be allocated to. If
\begin{eqnarray}
P({C_A}|x) > P({C_B}|x),
\end{eqnarray}
which can be indicated as the (20)
\begin{eqnarray}
\frac{{P({C_A})}}{{P(x)}}\mathop \Pi \limits_{i = 1}^d \mathop {P({x_i}|{C_A}) > }\limits_{} \frac{{P({C_B})}}{{P(x)}}\mathop \Pi \limits_{i = 1}^d \mathop {P({x_i}|{C_B})}\limits_{}.
\end{eqnarray}
The member is allocated to class ${C_A}$. $x$ and ${x_i}$ represent the attribute vector and the $i$th attribute in the vector. $d$ denotes the number of attributes.
\subsubsection{Step 4}
In our method, only one attribute is considered and it is the ${\varepsilon _i}$, residuals among the value of prediction and mean of history data. The $P(C)$ can be obtained by the (21)
\begin{eqnarray}
P({C_{_i}}) = \frac{{{N_i}}}{N}.
\end{eqnarray}
$N$ and ${N_i}$ represent the number of all the samples and the number of samples belong to class $P({C_i})$. Based on the history data, the ${N_i}$ can be obtained, if the ${\varepsilon _i}$ is least, the sample should be the class $P({C_i})$. The (20) can be converter as follow
\begin{eqnarray}
P({C_A})P({\varepsilon _A}|{C_A}) > P({C_B})P({\varepsilon _B}|{C_B}),
\end{eqnarray}
and $P({\varepsilon _A}|{C_A})$$\sim$$N({\mu _A},\sigma _A^2)$, $P({\varepsilon _B}|{C_B})$$\sim$$N({\mu _B},\sigma _B^2)$, $\mu$ and $\sigma$ are mean of the samples and standard deviation of the samples. Then, we can acquire the final equation. When
\begin{footnotesize}
\begin{eqnarray}
\frac{{P({C_A})}}{{\sqrt {2\pi } {\sigma _A}}}\exp ( - \frac{{{{({\varepsilon _{A,i}} - {\mu _A})}^2}}}{{2\sigma _A^2}}) > \frac{{P({C_B})}}{{\sqrt {2\pi } {\sigma _B}}}\exp ( - \frac{{{{({\varepsilon _B}_{,i} - {\mu _B})}^2}}}{{2\sigma _B^2}}),
\end{eqnarray}
\end{footnotesize}
it means in the next step, prediction results will be better from the model $A$, on the contrary, the model $B$ is the better choice. The hybrid output mode is named BARIMA in this paper.
\section{Performance Evaluation}
In this part, we apply the real data introduced in part II to verify and analyse our algorithm on the SPSS and Eviws platforms. The Gaussian distribution of fluctuant part, stationarity of the real data and reduction of error is also discussed. In order to make the result more convictive, data collected from two months ahead is used in the learning models to establish S-ARIMA, RARIMA and BARIMA models, and data of one week is used to test the relevant performance.
\subsection{Gaussian Distribution}
Based on the law of large numbers, the constant ${d_i}$ in six times can be calculated and applying K-S test to check the Gaussian distribution of residual ${\varepsilon _i}$. If the value of Statistical Significance is more than 0.1, the Gaussian distribution can be seen as correct. The result is shown in the Table I.
\begin{table}
\renewcommand{\arraystretch}{1.5}
 \caption{Kolmogorov-Smirnove Test for Six Data Sets}
\label{table_example}
\centering
\begin{tabular}{lcc}
\hline
Time &Sample Mean &Statistical Significance\\
\hline
03:00-07:00 &0 &0.412\\
07:00-11:00 &0 &0.611\\
11:00-15:00 &0 &0.609\\
15:00-19:00 &0 &0.266\\
19:00-23:00 &0 &0.612\\
23:00-03:00 &0 &0.951\\
\hline
\end{tabular}
\end{table}
From the it, the assumption that residual ${\varepsilon _i}$ with zero mean following Gaussian distribution is credible.
\subsection{Stationarity of Real Data}
In the last subsection, ${\varepsilon _i}$ is proven following stable Gaussian distribution. In order to use these models to forecast, these data sets should be stable. Based on the Eviews, we have checked the stationarity and the result is in the Table II. From the Table, all the absolute values of Test Statistic are greater than Test Critical Value 1\%level. It means all the data six sets are stable and can be used in the time series model directly. Relevant constant ${d_i}$ is also shown in the Table II.
\begin{table}
\renewcommand{\arraystretch}{1.5}
 \caption{Stationarity of Real Data}
\label{table_example}
\centering
\begin{tabular}{lccc}
\hline
Time &Test Critical Value 1\%level &Test Statistic &Constant\\
\hline
03:00-07:00 &-4.199 &-4.881 &244.865\\
07:00-11:00 &-4.199 &-5.318 &3334.73\\
11:00-15:00 &-4.199 &-5.270 &2412.30\\
15:00-19:00 &-4.199 &-6.166 &7718.78\\
19:00-23:00 &-4.199 &-6.425 &3427.81\\
23:00-03:00 &-4.199 &-4.782 &528.430\\
\hline
\end{tabular}
\end{table}
\subsection{Reduction of Error}
In this subsection, we apply the MAE (whose unit is number of people) on predicted results to estimate the performance of each model. Based on the last subsection, ${\varepsilon _i}$ and real data are stable, so we can take them into these models directly. At the first, the SARIMA and revised S-ARIMA are compared with each other to shown the transformation is applicable. The result is on the Table III.
\begin{table}
\renewcommand{\arraystretch}{1.5}
 \caption{Mean Absolute Error of SARIMA and S-ARIMA}
\label{table_example}
\centering
\begin{tabular}{lcc}
\hline
Time &SARIMA &S-ARIMA\\
\hline
03:00-07:00 &(2,0,2)(1,0,0$)_5$=15.09 &(1,5,7)=14.01\\
07:00-11:00 &(2,0,2)(1,0,0$)_5$=85.05 &(2,5,7)=79.82\\
11:00-15:00 &(3,0,3)(1,0,0$)_5$=162.11 &(3,5,9)=140.62\\
15:00-19:00 &(3,0,2)(1,0,0$)_5$=159.40 &(5,8)=131.98\\
19:00-23:00 &(2,0,2)(1,0,0$)_5$=129.04 &(2,5)=117.04\\
23:00-03:00 &(1,0,1)(1,0,0$)_5$=57.24 &(1,4,5)=66.42\\
\hline
\end{tabular}
\end{table}
From the table we can see the transformation of the SARIMA model is effective to reduce error. It may reduce about 7\% of MAE of SARIMA. So, the following works is based on the revised S-ARIMA model.

Based on the (17) and the constant ${d_i}$ in the Table II, the complete RARIMA model can be obtained. In particularly, the RARIMA model should be revised from the S-ARIMA rather than SARIMA. From the Table IV, the predicted results can be compared with each other.
\begin{table}
\renewcommand{\arraystretch}{1.5}
 \caption{Mean Absolute Error of S-ARIMA and RARIMA}
\label{table_example}
\centering
\begin{tabular}{lcc}
\hline
Time &S-ARIMA &RARIMA\\
\hline
03:00-07:00 &(1,5,7)=14.01 &(1,5,7)=12.13\\
07:00-11:00 &(2,5,7)=79.82 &(2,5,7)=66.09\\
11:00-15:00 &(3,5,9)=140.62 &(3,5,9)=125.78\\
15:00-19:00 &(5,8)=131.98 &(5,8)=142.95\\
19:00-23:00 &(2,5)=117.04 &(2,5)=113.77\\
23:00-03:00 &(1,4,5)=66.42 &(1,4,5)=47.3\\
\hline
\end{tabular}
\end{table}
From the Table IV we can seen the RARIMA model is better than S-ARIMA model in most situations. Now, we analyse the two model from a picture that describes the predicted results of them in 19:00-23:00.
\begin{figure}
  \centering
  \includegraphics[width=3.5in]{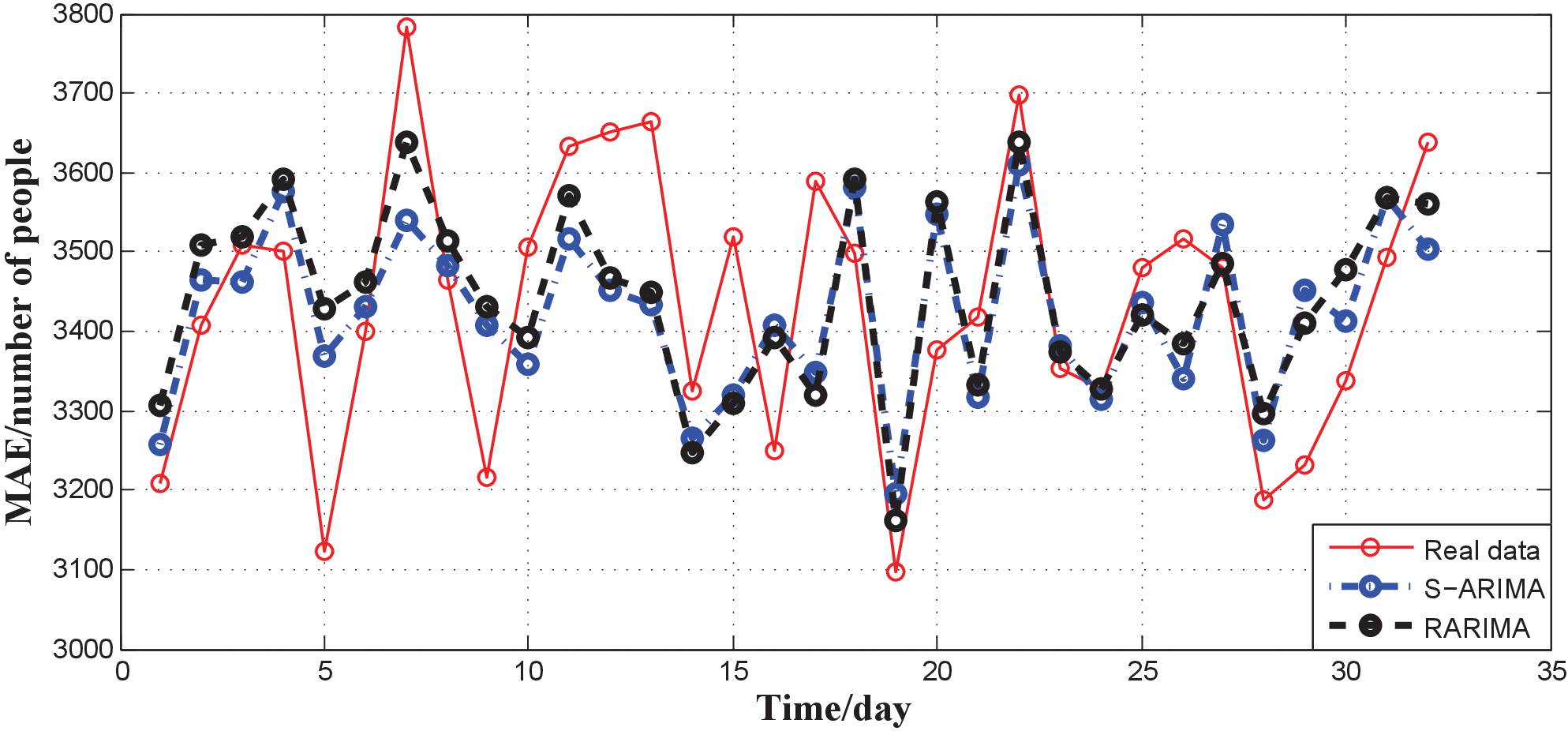}\\
  \caption{Predicted Results of Two Models in 19:00-23:00}
\end{figure}
From the Fig. 2, we can see the RARIMA is better to track the real data, especially for the peak. however, the S-ARIMA is more stable, when the real data is increasing and decreasing placidly. So, if we use the RARIMA model to track peak and use the S-ARIMA to track other situation, results are better. Inspiration for constructing a hybrid model is derived from the analysis and detail theories have been explained in the Model Optimization in part II.

We use the Bayesian learning algorithm to choose the best model in the next prediction. Corresponding results are shown in the Table V.
\begin{table}
\renewcommand{\arraystretch}{1.5}
 \caption{Performance of BARIMA}
\label{table_example}
\centering
\begin{tabular}{lcccc}
\hline
Time &SARIMA &RARIMA &BARIMA &Minimum error\\
\hline
03:00-07:00 &15.09 &12.13 &11.25 &9.08\\
07:00-11:00 &85.05 &66.09 &69.06 &42.05\\
11:00-15:00 &140.62 &125.78 &112.65 &86.36\\
15:00-19:00 &162.11 &142.95 &124.53 &87.44\\
19:00-23:00 &129.04 &110.42 &113.77 &99.75\\
23:00-03:00 &57.24 &47.30 &45.79 &37.33\\
\hline
\end{tabular}
\end{table}
From the table we can see compared with SARIMA, performance of the hybrid model BARIMA is obviously outstanding. It reduce the 20\% MAE of the SARIMA. In order to explain the improvement, the Fig. 3 is shown as follow and a minimum error of BARIMA is provided to analyze our model thoroughly.
\begin{figure}
  \centering
  \includegraphics[width=3.4in]{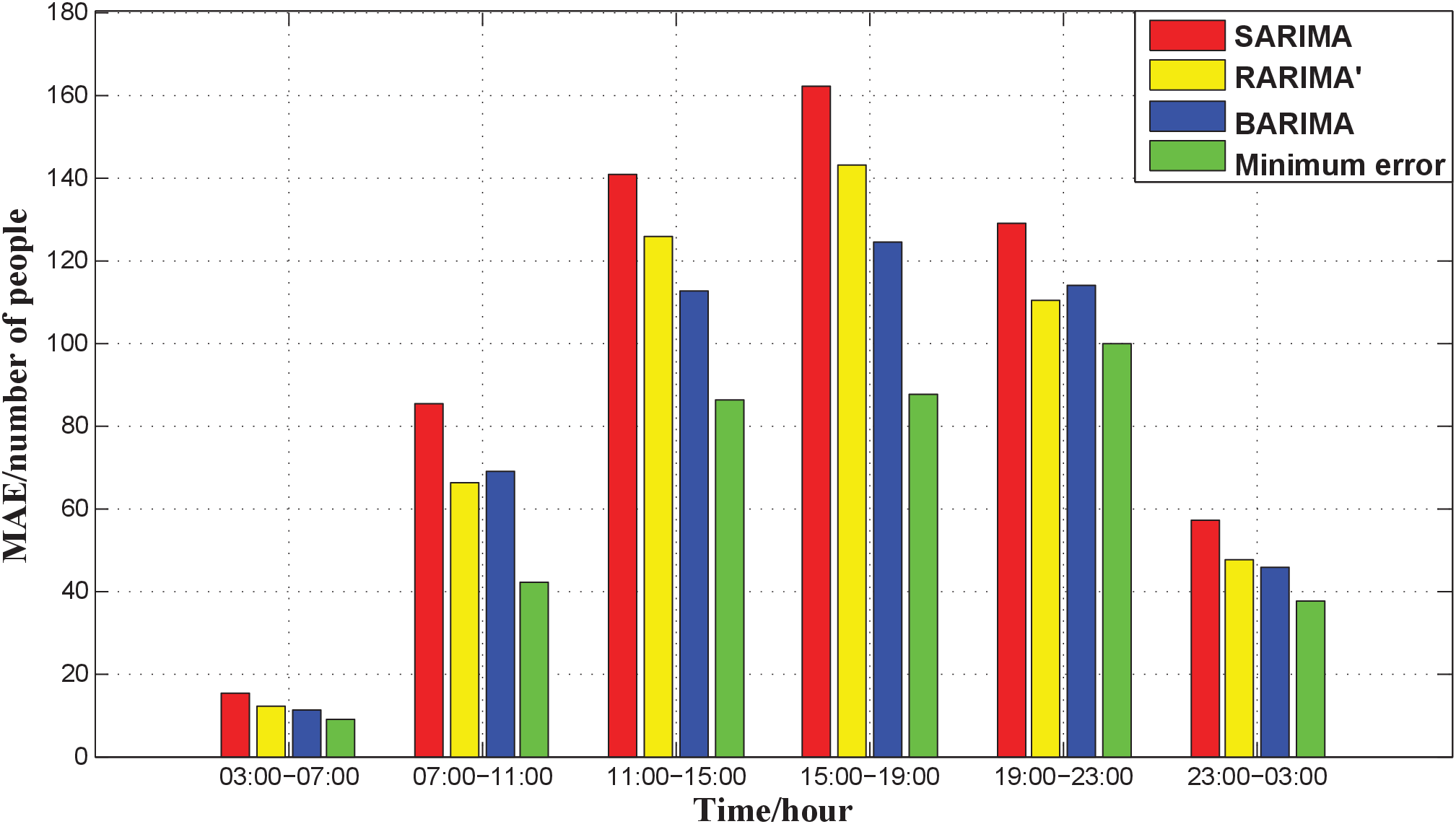}\\
  \caption{Performance of BARIMA}
\end{figure}

From the Fig. 3, performance of BARIMA is better than SARIMA. Combining two model to generate a hybrid model model is a correct method to improve the accuracy. Nevertheless, it is hard to exactly choose the best model to forecast each time. In our paper, the Bayesian learning algorithm is effective in most time. When the $P({\varepsilon _A}|{C_A})$ is similar with $P({\varepsilon _B}|{C_B})$, the algorithm can not work well such as at 07:00-11:00 and 19:00-23:00. The major reason is that only one attribute is used in the Bayesian learning algorithm.
\subsection{Performance Test of BARIMA}
Based on the training sets, training result have shown the BARIMA model can reduce the prediction error of SARIMA. However, it is not reliable only from training result to evaluate anyone model. The more professional method is that using the complete model obtained from training sets to predict remaining data and test the performance of these models. In this paper, data of one week after the two months is used to finish the task. In order to compare our model with some models in \cite{13}, the MAE is normalized and the the mean absolute percentage error (MAPE) is used to estimated each model. The MAPE about minimum error of BARIMA is also provided. Detailed information is shown in Table VI.
\begin{table}[!h]
\renewcommand{\arraystretch}{1.5}
 \caption{MAPE of Prediction}
\label{table_example}
\centering
\begin{tabular}{lccccc}
\hline
Model &SARIMA &SM &RW &ANN &BARIMA\\
\hline
MAPE(\%) &6.28 &6.76 &6.30 &5.80 &4.93\\

\hline
\end{tabular}
\end{table}
From the testing result, the BARIMA model performs better than the
classical SARIMA, RW and SM, where RW is a simple baseline that predicts traffic in the
future that is equivalent to current conditions, and SM predicts the average
in the training set for a given time of the day. In addition, the BARIMA can also be improved by prefect Bayesian learning algorithm to obtain the minimum error.
\section{Conclusion and Future Works}
In this paper, by analyzing traffic flow features and structure of SARIMA model, we propose a hybrid model to improve accuracy of short-time traffic forecasting. Firstly, based on the autocorrelation of traffic follow, classical SARIMA is revised to prevent irrelevant data from predicting traffic flow. Secondly, the traffic flow is divided into fluctuant and stabile parts to reduce the content of prediction. Finally, according to Gaussian distribution of residuals, a Bayesian learning algorithm is applied to conduct our proposed scheme, i.e., BARIMA. Extensive simulation results show that proposed scheme performs better in comparison with existing schemes.

In the future, we will combine abundant non-transport data sets to perfect proposed Bayesian learning algorithm, and attempt to improve the performance of our scheme.



\end{document}